%% file: main.tex
\documentclass[11pt]{article}
\usepackage[margin=1in]{geometry}
\usepackage{amsmath,amssymb}
\usepackage{booktabs}
\usepackage{graphicx}
\usepackage{pgfplots}
\pgfplotsset{compat=1.18}
\usepackage[numbers,sort&compress]{natbib}
\usepackage[hidelinks]{hyperref}
\usepackage{kotex}

\date{September 11, 2026}

\title{Metric--Construction Coupling Inflates\\Measured Synthetic Dialect Recovery}

\author{Hyojung Han\\ThakiCloud\\\texttt{hyojung.han@thakicloud.com}}

\begin{document}
\maketitle
\bibliographystyle{plainnat}

\begin{abstract}
Korean dialect corpora are available but not redistributable: weights may be released,
while reproducing training and evaluation from the underlying data cannot be. We ask how
much of that supervision synthetic data recovers under a matched base model and serving
configuration, and whether that recovery can be measured independently of the synthesis
pipeline. We contribute \textbf{KoDialectBench}, 1{,}000 items across five regions on three
axes, released as identifier hashes and scoring code so users reconstruct the items from
their own licensed copy. Recovery is strongly axis-dependent: our best synthetic arm reaches
91.2\% of the real-data gain on region identification but 63.7\% on comprehension. On
generation the answer depends on the metric: the \emph{deployed} marker lexicon reports 92.3\%
recovery on dialectness and 119.3\% on region match --- the latter exceeding the real-data
reference --- whereas reference-based generation reaches 72.4\%. We find the marker metrics' scoring
inventory is entirely contained in the inventory our transformation rules can emit. We test the
effect of construction access directly with an \emph{exact-form} construction-disjoint arm that
withholds 20\% of marker types from the transformation rules; residual suffix-trigger overlap
remains in 239 of its 8{,}600 training targets (2.78\%), which we report rather than claim
away. \textbf{At
exactly matched training size} (8{,}600 examples, matched task and region composition, source
register and recipe) it reduces dialectness recovery from 91.8\% to 8.1\% and region-match
recovery from 101.9\% to 25.6\% \emph{on the held-out marker inventory} --- the same scorer as
above with the inventory swapped, which is why the coupled arm's figure differs --- while the three pipeline-independent measurements do not fall at
all, moving by $+3.3$, $+2.0$ and $+2.3$ recovery points --- no larger than the corresponding
observed SK5--SK6 differences. A complementary evaluator sweep defines
\textbf{metric--construction coverage} (MCC) as the fraction of the evaluation inventory
contained in the emittable inventory, and finds measured \emph{dialectness} recovery increasing monotonically as
overlap rises from $\mathrm{MCC}=0$ to $\mathrm{MCC}=1$, with non-overlapping 2.5--97.5
percentile ranges across sampled evaluator inventories at the endpoints. Shared construction and evaluation inventories can
therefore substantially inflate estimates of synthetic-data recovery.
\end{abstract}

\section{Introduction}

Dialect NLP is treated as a low-resource problem. For Korean it is more precisely a
\emph{licensing} problem. AI Hub publishes conversational corpora for five
regions~\citep{aihub_dialect}, but redistribution and export are restricted. Model
weights derived from them may be released; what cannot be released is the data needed
for an independent party to reproduce the training or the evaluation.

Rule-based synthesis is a common response. VALUE and
Multi-VALUE~\citep{ziems2022value,ziems2023multivalue} construct English dialect
perturbations from attested feature inventories, and DADA~\citep{liu2023dada} composes
adapters over such rules. What is seldom reported is \emph{how much of the real-data
gain synthesis recovers}, measured on one benchmark, from one base model, under one
serving configuration, and separated by ability.

Our central finding is not the recovery figure itself but its structure. Recovery
differs sharply by axis, and on the generation axis the answer depends on whether the
evaluation resource is independent of the generation pipeline. When it is not,
synthetic supervision appears to \emph{surpass} the data it was meant to substitute
for. We treat that as a measurement failure rather than a result, and we show how to
detect it.

\paragraph{Research questions.}
\textbf{RQ1.} How much does synthetic dialect supervision recover under
construction-independent evaluation? \textbf{RQ2.} How does metric--construction coupling
bias the estimated recovery? We also report one secondary diagnostic: a marker-based over-separation pattern does not
survive a marker-free probe (Appendices~\ref{app:probe} and~\ref{app:oversep}).

\paragraph{Contributions.}
(i)~KoDialectBench, a three-axis Korean dialect benchmark released under a
licence-compatible reconstruction scheme.
(ii)~A recovery-rate protocol with a matched control, together with the two control
failures we had to fix before any number was meaningful.
(iii)~A demonstration that a lexicon-based dialect metric becomes confounded with
construction access when its scoring inventory is contained in the synthesis inventory,
together with a disjointness test that detects this condition.
(iv)~\textbf{MCC}, a coverage statistic for the coupling, together with a sweep showing
that measured recovery rises with it and that the real-versus-base separation margin
shrinks as it drops.
(v)~A construction-time intervention --- splitting the marker inventory before synthesis so
the rules and the scorer share no exact form, with the residual suffix-trigger overlap
quantified separately --- which reduces marker-based recoveries by 76--84 percentage points
while leaving every pipeline-independent measurement within the observed SK5--SK6 seed
difference.

\section{Related work}

\paragraph{Dialect modelling and synthetic dialect transformation.} Dialect NLP has long
modelled variation directly, from demographic dialect variation in social
media~\citep{blodgett2016aae} to dialect behaviour in generation~\citep{groenwold2020aave} and
the dialect discrimination reproduced by current LLMs~\citep{fleisig2024bias}. Rule-based synthesis
is a common response to data scarcity: VALUE~\citep{ziems2022value} quantified dialect
disparity in NLU by transforming Standard American English with attested morphosyntactic
features, Multi-VALUE~\citep{ziems2023multivalue} extended this across many English varieties
with eWAVE attestation tiers, DADA~\citep{liu2023dada} learned to compose feature-specific
adapters, and EnDive~\citep{gupta2025endive} pairs synthetic dialect transformation with human
validation and downstream benchmarking. Parallel standard/dialect benchmarks now exist for
reasoning~\citep{lin2025redial} and for judge consistency across
varieties~\citep{faisal2025toxicity}, and LoRA-based dialect adaptation has been studied
directly~\citep{maheshwari2026lora}. Our transformation stage follows this lineage, mapping
marker purity onto tiers analogous to attestation; the differences in setting are that our
target varieties are Korean regional dialects and that our motivation is a licence constraint
rather than robustness alone.

For Korean specifically, automatic dialect identification has been pursued on the speech
side through intonation modelling~\citep{lee2021intonation}, prosodic and segmental
ensembles for forensic profiling~\citep{lee2022forensic}, and self-supervised speech
representations~\citep{lee2025wav2vec}, the last of which also questions whether
administrative provinces are the right class boundary --- a caveat that applies to our own
five-region scheme. \textbf{Closest to our setting}, \citet{park2026korean}
propose an iterative refinement framework for faithful standard-to-dialect Korean
translation and identify an evaluation distortion of their own: $n$-gram metrics reward
copying the source, so they introduce dialect-fidelity and target-dialect-ratio measures
instead. Low-resource dialect fine-tuning has also been studied for Lebanese
Arabic~\citep{yakhni2025lebanese}. \textbf{Our failure mode is a different one} --- not a
metric that rewards copying, but \emph{feature-inventory coupling} between the synthesis
rules and the scorer --- and we manipulate that coupling directly, through
construction-disjoint training and an overlap sweep, rather than proposing a replacement
metric.

\paragraph{Evaluation validity for controlled generation.} Our claim is a construct-validity
claim: a score can be a valid operationalisation of something other than the construct it is
read as. Measurement modelling imports this distinction from the social sciences into machine
learning~\citep{jacobs2021measurement}, and applying it to NLP benchmark datasets shows that
the dataset's own construction choices can determine what the score
tracks~\citep{blodgett2021salmon}. Style transfer, the closest task family to ours, has an
established meta-evaluation literature reaching the same conclusion from the metric side:
style strength, content preservation and fluency must be measured
separately~\citep{mir2019evaluating}; automatic metrics are interdependent and manipulable,
so reported gains need reruns and error bars~\citep{tikhonov2019retrain}; metric--human
correlation is weak and task-dependent~\citep{briakou2021evaluating,lai2022compass}; and the
field's validation practice is inconsistent enough to warrant
standardisation~\citep{ostheimer2023call,ostheimer2024llm,schmidtova2024metrics,mukherjee2025reliable}.
Most directly, \citet{pauli2025stylegap} show that properties of
the \emph{test data} can inflate a metric's apparent validity. Our contribution is adjacent
but distinct: rather than meta-evaluating metrics against human judgement, we manipulate the
containment relation between the synthesis inventory and the scoring inventory and measure
the dose--response. Goodhart-style proxy failures have been categorised by
mechanism~\citep{manheim2018goodhart}; our setting is Goodhart-like in that a proxy ceases to
track the intended construct once the data-construction process has privileged access to the
features the proxy rewards. \textbf{We do not assign it to a specific Goodhart subtype}: the
marker metric is not itself the training objective, so neither the regressional nor the causal
variant applies cleanly. We did not run a human evaluation, a limitation to read against the
reporting standards this literature
recommends~\citep{vanderlee2019best,howcroft2020confusion}.

\paragraph{Evaluating synthetic training data.} Synthetic data generation, curation and
evaluation now has its own survey~\citep{long2024survey}, and a recurring finding is that
intrinsic quality of the generated text and downstream usefulness come
apart~\citep{iskander2024quality}; generator quality can therefore also be assessed through the models
trained on their outputs~\citep{kim2025generators}, and style, meaning, and divergence are better reported
separately than fused~\citep{chim2025synthetic}, motivating our separation of the three evaluation axes. For
low-resource languages specifically, the real-versus-synthetic downstream gap has been
measured directly~\citep{anikina2025lowresource}, which is the closest prior work to our
recovery question; the nearest established primitive is train-on-synthetic /
test-on-real~\citep{esteban2017rcgan}, which compares a synthetic-trained model against real
data but does not express the result as a fraction of a real-data arm's gain over an
untrained baseline. \textbf{We are not aware of a prior paper that defines such a recovery
rate as its object of study}; we state this as a gap rather than claim priority over a
literature we may have missed. A separate line shows that training on generated text reduces
linguistic diversity~\citep{guo2023diversity} and that recursive training degrades the tails of
the distribution~\citep{shumailov2024collapse,alemohammad2024mad}. That is related evidence
that synthetic training can narrow distributional support, but it concerns recursive
model-generated data; \textbf{the lexical-inventory ceiling we describe is a distinct
phenomenon} that those papers do not test.

\paragraph{Benchmark artifacts, contamination, and construction coupling.} A benchmark's
construction can create predictive structure nobody intended: annotation artifacts let a
model score well from the hypothesis alone~\citep{gururangan2018artifacts,poliak2018hypothesis},
and aggregate accuracy can diverge from the capability it is read as~\citep{ribeiro2020checklist}.
Contamination is the better-known modern case, in which evaluation items appear in training
data~\citep{sainz2023trouble,sainz2024conda}; it has been reported at
scale~\citep{li2024contamreport}, shown to distort generalisation
estimates~\citep{dong2024generalization}, and attacked by decontaminating at inference
time~\citep{zhu2024decontamination}. Synthetic evaluation sets carry their own version of the
problem --- they can be formally valid yet easier than human-written
benchmarks~\citep{gill2025lost} --- and practitioners building judge workflows already name
circularity as a risk~\citep{cooper2025evalassist}. Automatic benchmarks can also be inflated
with no change in underlying behaviour at all~\citep{hsia2024goodhart}. \textbf{We separate
three failures.} Contamination is exposure of evaluation \emph{items}; annotation artifacts
are unintended regularities in a dataset; the coupling we report is neither --- the
\emph{feature inventory} of the evaluation is shared with the pipeline that produced the
training data, so no item leaks, no annotator introduced a shortcut, and the metric is still
inflated. For rule-based dialect synthesis that overlap can be total, which is what we find.

\section{KoDialectBench}

\subsection{Composition}

The benchmark holds 1{,}000 items, 200 for each of Gangwon, Gyeongsang, Jeolla,
Chungcheong, and Jeju, scored on three tasks. \textbf{T1 comprehension} maps a dialect
sentence to standard Korean, scored by chrF~\citep{popovic2015chrf} and exact match.
\textbf{T2 identification} is a five-way region choice, scored by accuracy.
\textbf{T3 generation} maps a standard sentence and a target region to that region's
dialect. We report two families of T3 score and keep them separate throughout:
a \emph{marker-lexicon} family (dialectness, region match, and their product ADI2) and
a \emph{reference} family (chrF against the human dialect sentence paired with the
input). Section~\ref{sec:coupling} explains why the distinction is load-bearing.

We define the marker-lexicon family as a \textbf{lexical dialect-signal metric}, not a
quality score. It counts how many words carry known regional markers. It does not
measure whether the output is a natural utterance.

\subsection{Establishing discriminative power}

Before comparing arms we checked that the marker metric separates known-different
inputs. On a held-out split of 2{,}000 pairs (400 per region), pairwise separation of a dialect sentence
from its paired standard sentence is 96.5\% (Jeolla), 92.5\% (Gangwon), 90.5\%
(Chungcheong), 86.8\% (Gyeongsang), and 86.5\% (Jeju). Own-region score minus the best
other-region score is positive in all five regions, between $+0.56$ and $+0.78$. This
establishes that the metric is not inert; it does not establish construct validity for
generation quality, and \S\ref{sec:coupling} shows it does not establish independence
from our synthesis pipeline either.

\subsection{Release under a restrictive licence}

\textbf{The provider has confirmed the boundary in writing.} Models and weights produced by
actually training on the corpus are derivative works that may be released without restriction
on purpose; the original data, and anything that merely extracts, edits, converts or
re-processes it, may not be redistributed --- including storage on overseas platforms, which
the provider treats as export. Weights may be released only if such data is not bundled with
them, and any release must name the datasets and credit the provider. \textbf{The
dialect-to-standard correspondence inventory falls on the forbidden side}: it is extracted
from the corpus's own token-level labels, not learned by a model, so we cannot release it.

Our artifact is built to that boundary. We distribute one-way hashes of item identifiers, a
digest of each item body for reconstruction verification, the scoring code executed in
isolation with no repository on the interpreter path, and a script that regenerates the marker
lexicon locally with a pinned checksum. A user reconstructs the same 1{,}000 items from their
own licensed copy. The artifact contains neither dialect text nor the marker lexicon: the
1{,}000 released records carry identifiers, region, digests and length statistics only.

An eight-axis gate enforces this. One axis executes only copies of the released files
in a temporary directory. Our first implementation of that axis imported the scoring
code through the interpreter path and therefore passed while still depending on the
repository; a negative test exposed it. We report this because a self-satisfying
artifact gate is an easy failure to ship.

\section{Experimental setup}

\subsection{Arms and training}\label{sec:arms}

All arms start from the same base model (Qwen3.8-27B, bf16) and use identical LoRA
settings: rank 8, $\alpha=16$, all linear projections, learning rate
$2\times10^{-5}$, one epoch, adapters merged before serving. \textbf{SK10-M is the one
exception}: it stops at a fixed 8{,}590 updates rather than completing an epoch
(\S\ref{sec:disjoint}). \textbf{arm R} trains on
8{,}600 real parallel examples and serves as the real-data reference arm. \textbf{arm C} and
\textbf{SK5} train on 8{,}600 synthetic examples, differing in source register;
\textbf{SK12-D9b} is the exactly size-matched 8{,}600-example construction-disjoint
intervention and the primary comparison of \S\ref{sec:disjoint}; \textbf{SK11-D9} (8{,}758)
is an independently sampled disjoint arm and \textbf{SK10-M} a fixed-update control;
\textbf{SK7-D} and \textbf{SK9-D} form a 26{,}861-example disjoint seed pair used only for
sensitivity analysis;
\textbf{SK4} trains on 26{,}600 synthetic examples; \textbf{SK6} repeats SK5's
configuration under a different seed. All training sets mix the three tasks evenly and
sample regions uniformly for T2, because the base model has a strong prior toward two
regions and an unbalanced set amplifies it. Full hyperparameters, prompt templates,
generation sampling parameters, and serving flags are released with the artifact (one
template discrepancy is disclosed in Appendix~\ref{app:controls}).

\subsection{Two preconditions we had to repair first}

We verified two preconditions before computing any recovery figure. First, the base and
reference arms must differ only in the intended supervision: our initial baseline was a
quantised model from a different family, which moved base model, numeric precision and
training data together, and recovery became interpretable only after re-measuring a plain
bf16 base. Second, a deployed adapter must actually be active: serving the LoRA directly
was silently ignored on this stack, with token log-probabilities identical to five decimal
places between base and the supposedly adapted endpoint. We now merge every adapter before
serving and gate on the adapted endpoint differing from base. Details and the negative
tests are in Appendix~\ref{app:controls}. The general form is worth stating: before asking
how much synthetic supervision recovers, verify that the arms differ only in the intended
treatment and that the deployed adapter is active.

Greedy decoding is not reproducible in practice on this stack, so every cited number
is the median of three repeats under one fixed serving configuration.

\section{The synthetic pipeline}\label{sec:pipeline}

Region markers are mined as (form, frequency, purity) triples from the training split
only; the benchmark test split contributes to neither the lexicon nor the purity
statistics. Purity maps to an attestation tier ($\geq0.95 \to 1.0$,
$\geq0.85 \to 0.6$, else $0.3$), following Multi-VALUE's use of eWAVE
tiers~\citep{ziems2023multivalue}. Per-sentence transformation density is sampled from
the observed real distribution rather than fixed, since a fixed intensity yields a
distribution more uniform than reality.

\paragraph{Refining the rules did not move recovery.} Three transformation-side
refinements --- preserving marker frequency when sampling variants, mapping purity to
attestation tiers, and sampling density --- produced no appreciable change.

\paragraph{Source register strongly affected rule applicability.} Markers were mined
from spoken conversation, while the sentences being transformed were written-register
prose generated by a 27B model, so rule applicability ran at 22--62\% of the real rate,
with Jeolla at 5.4\% against 23.2\%. A small preflight over prompt variants raised
applicability from 39.0\% of the real rate (written prose) to 79.6\% (an explicit
colloquial-utterance instruction). A third variant that steered the model toward dialect
vocabulary directly performed worse than the colloquial instruction; we do not report a
figure for it because the value was not preserved in the committed artifact.

\textbf{This roughly twofold increase in applicability did not translate into a
comparable improvement on the pipeline-independent metrics.} Comparing arm C (written
source) with SK5 (colloquial source) at equal training-set size, the differences are
$-0.49$ chrF on comprehension, $+0.60$\,pp on identification, and $+0.43$ chrF on
reference-based generation --- 0.3, 0.5 and 0.5 times the observed seed-to-seed difference of
\S\ref{sec:seed} respectively. Arm C's full row is in Table~\ref{tab:full}. Rule coverage and downstream capability recovery are
therefore separate axes, and an applicability gain should not be reported as a recovery
gain.

\paragraph{Training-set size, under a fixed recipe.} Quadrupling the source pool
produced little additional rule coverage in earlier exploratory analysis, while growing the
training set produced more. Those counts predate the structured ledger and are not
reproducible from the released artifacts, so we do not quote them and no recovery figure
depends on them. We state the downstream claim in the form the experiment supports:
\emph{under our fixed one-epoch recipe, increasing the number of synthetic training
examples from 8{,}600 to 26{,}600 improved region identification by 9.9 times the observed
SK5--SK6
variation, while comprehension and reference-based generation moved within it.} Because
every arm trains for one epoch, the 26{,}600-example arm also receives roughly three times the
optimizer updates, so data quantity and optimization compute are confounded; we do not
claim training-set size alone as the cause.

\paragraph{A coverage limit that probability tuning does not cross.} Raising the
transformation probability to $p{=}1.0$ does not bring synthetic dialectness to the real
level, because the limit is the size of the marker inventory rather than how often it fires.
Our earlier per-region estimates of that gap predate the structured ledger, so we do not
quote them; \S\ref{sec:disjoint} measures the same limitation directly and with ledger
support, by removing a fifth of the inventory and observing held-out dialectness recovery of
8.1\%.

\section{Results}

\subsection{Recovery differs by axis}\label{sec:axis}

Recovery is $(\text{arm}-\text{base})/(\text{arm R}-\text{base})$ on medians of three
repeats (Table~\ref{tab:main}). We report it \textbf{per axis} rather than as a single
aggregate. A pooled median over the individual metrics would give the generation axis
three of five votes, one of which (ADI2) is the product of the other two, and would
therefore report generation performance under the name of overall recovery. Where a
single number is unavoidable we use the axis-balanced mean.

\begin{table}[t]
\centering\small
\input{tables/tab-main.tex}
\caption{KoDialectBench v0, medians of three repeats, one serving configuration.
$^\dagger$ marker-lexicon family --- see \S\ref{sec:coupling} before interpreting.
$^\ddagger$ guardrail, not a gain axis (\S\ref{sec:guardrail}). SK6 repeats SK5 with a
different seed.}
\label{tab:main}
\end{table}

\begin{figure}[t]\centering
\includegraphics[width=3.4in]{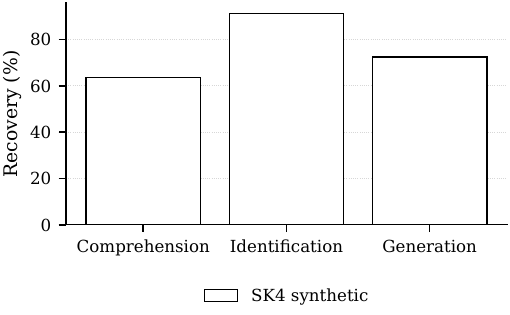}
\caption{Capability recovery of the best synthetic arm (SK4, 26{,}600 examples) under
construction-independent evaluation: chrF for comprehension, accuracy for identification,
and reference chrF for generation. Intermediate arms are in Table~\ref{tab:main}; the
coupled generation metric is treated separately in Figure~\ref{fig:coupling}.}\label{fig:axis}
\end{figure}

For SK4 the axis vector is \textbf{63.7\% comprehension, 91.2\% identification}, and
for generation \textbf{72.4\% by reference chrF}. The marker-based metrics disagree,
reporting 92.3\% on dialectness and 119.3\% on region match --- the latter above the
real-data reference (Figure~\ref{fig:axis}). \textbf{We report the marker metrics
separately and do not average them}: ADI2 is the product of dialectness and region match,
so a mean over the three counts the same two quantities twice. Taking the reference figure
for generation, the axis-balanced mean is 75.8\%.

\subsection{Metric--construction coupling}\label{sec:coupling}

The marker-lexicon family reports SK4 above the real-data reference on region match
(119.3\% recovery) and ADI2 (105.2\%). Before treating that as a result we asked
whether the evaluation resource is independent of the generation pipeline.

It is not. We enumerated every dialect form the transformation rules can emit --- the
union of all lexical variants and all suffix variants they may apply --- and removed
those forms from the scoring lexicon. \textbf{The held-out remainder is empty in all
five regions} (Table~\ref{tab:disjoint}). The scoring lexicon is a subset of what the
transformer can produce, so the metric cannot distinguish producing genuine dialect
from producing the forms our rules happen to know.

\begin{table}[h]\centering\small
\input{tables/tab-disjoint.tex}
\caption{Disjointness test. No scoring marker survives removal of the forms the
transformation rules can emit.}\label{tab:disjoint}
\end{table}

\begin{figure}[t]\centering
\includegraphics[width=3.4in]{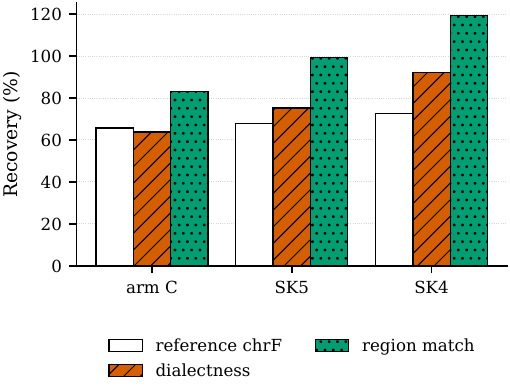}
\caption{Generation recovery under an independent metric (chrF against held-out human
references) and the two coupled marker metrics \emph{as deployed} (full scoring lexicon;
Table~\ref{tab:disjoint-result} instead scores the held-out 20\%), whose inventory the transformation rules can
emit. The two marker metrics are shown separately rather than averaged: ADI2, the third
marker metric, is their product. The ordering of arms is preserved; the level, and the
crossing of the 100\% line, is not.}\label{fig:coupling}
\end{figure}

We therefore rescored generation with chrF against the held-out human dialect
reference, a metric that does not use the transformation-derived marker lexicon. We claim
independence of the \emph{scoring inventory}, not lexical disjointness: a human reference
naturally shares ordinary Korean vocabulary with anything our pipeline produces. Generation
recovery is \textbf{72.4\%} on this metric, against 92.3\% and 119.3\% on the two marker
metrics, and no synthetic arm exceeds the real-data reference
(Figure~\ref{fig:coupling}). The apparent overshoot was metric alignment.

\subsection{Intervening at construction time}\label{sec:disjoint}

Section~\ref{sec:coupling} establishes that the two inventories coincide. To test the effect
of construction access directly, we intervene at synthesis time, splitting the
marker inventory \emph{before} synthesis: 80\% by stratified sample (region, frequency
quartile, purity quartile) drives the transformation rules, and the held-out 20\% is used
only for scoring. Before the cross-region emission ban is applied, the region-wise construction and evaluation
partitions overlap once their global unions are taken (240 eojeols and 7 suffixes), which
matters because scoring fires across regions. The global ban removes the eojeol forms from
the emittable inventory; it does not ban suffix entries, so one of the 7 overlapping suffixes
(a Chungcheong ending) stayed emittable and the post-ban exact-form intersection is 0 eojeols
and 1 suffix; we report rather than repair this. We then filtered the training set until no
generated example carried an evaluation marker its source did not already contain.

\textbf{We call this construction-disjoint, not training-unseen.} The full pool holds 26{,}861
examples; SK11-D9 draws 8{,}758 from it and SK12-D9b 8{,}600, and the counts below are for
the data each model actually saw. In SK11-D9, four targets (0.05\%) and 39 inputs (0.45\%)
contain an evaluation form exactly; in SK12-D9b, three targets (0.03\%) do. All are forms
already present in the source sentence.

\textbf{Exact-form disjointness is not the same as scoring disjointness}, and we report the
gap rather than leave it implicit. The evaluator fires on exact eojeol match \emph{or} suffix
match, while the ban list blocks exact forms; a form can therefore be construction-disjoint by
exact match and still trigger the evaluator through its ending. Counted with the scorer's own
firing rule, \textbf{this happens in 239 of SK12-D9b's 8{,}600 targets (2.78\%)} and 234 of
SK11-D9's 8{,}758 (2.67\%); a further 3 and 4 targets respectively contain an exact evaluation
form and are counted separately. \textbf{The direct contamination channel favours higher rather than
lower held-out marker scores}, which makes the observed collapse conservative with respect to
this leakage mechanism --- though training is non-linear, so we do not claim a guaranteed
direction for the final model's score.

\textbf{The evaluator separates before we interpret anything.} On the held-out marker
inventory the real-data arm sits well above base --- dialectness 0.0092 to 0.0387 ($+321$\%
relative) and region match 0.0280 to 0.2345 ($+738$\%). To confirm that the recovery ratios
below are not a division by noise, we bootstrap the denominator itself: pairing arms on item
index and resampling the 1{,}000 benchmark items 2{,}000 times gives an $\text{arm R}-\text{base}$
gap of $+0.0292$, CI $[+0.0250, +0.0332]$ on dialectness and $+0.2040$, CI
$[+0.1785, +0.2303]$ on region match. Both exclude zero by a wide margin. (These point estimates
differ from the figures above --- in the fourth decimal for dialectness, the third for region
match --- because the bootstrap pairs items within a single run (the second repeat), while the
figures above are medians across three runs.)

\begin{figure}[t]\centering
\includegraphics[width=3.4in]{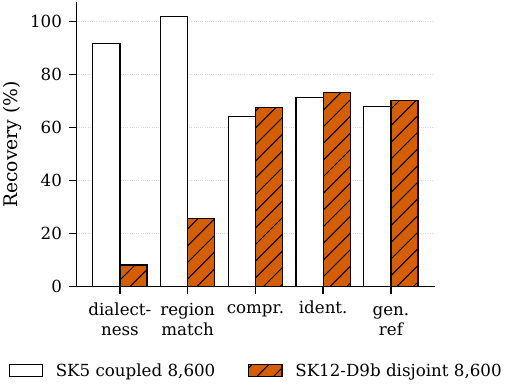}
\caption{The same training-side intervention, read by two metric families. Marker columns are
scored on the \emph{held-out} 20\% inventory (not the deployed lexicon of
Figure~\ref{fig:coupling}). Restricting the transformation rules to 80\% of the marker
inventory collapses the two marker-based measurements while the three pipeline-independent
ones do not fall at all.}\label{fig:disjoint}
\end{figure}

\begin{table}[t]\centering\small
\input{tables/tab-disjoint-result.tex}
\caption{Construction-time intervention. \textbf{SK12-D9b is the primary comparison}: it is
exactly size-matched to SK5 (8{,}600 examples) and matches it on task-by-region composition,
source register and optimization recipe; 66.2\% of its source sentences are shared with SK5.
SK11-D9 is an independently sampled disjoint arm at 8{,}758, reported because the effect
reproduces across two different synthetic samples. The treatment in both is a
construction-disjoint synthesis pipeline: 20\% of marker types are withheld from the transformation inventory, and residual
evaluation-form leakage is filtered. \textbf{The marker columns are scored by the same
scorer code as Table~\ref{tab:main} but on the held-out 20\% inventory} instead of the
deployed lexicon, so the same arm carries a different marker figure in the two tables.
Pipeline-independent columns report comprehension chrF, five-way region-identification
accuracy, and generation chrF against held-out human references.}\label{tab:disjoint-result}
\end{table}

\textbf{The intervention produces a 76--84 point reduction on the marker metrics without a
corresponding decrease on the three pipeline-independent measurements}
(Table~\ref{tab:disjoint-result}). Our primary
comparison is SK5 against \textbf{SK12-D9b}, which is exactly size-matched at 8{,}600
examples. On the held-out marker evaluator, removing a fifth of the inventory from the rules
collapses the score: dialectness recovery falls from 91.8\% to \textbf{8.1\%} and region
match from 101.9\% to \textbf{25.6\%}, gaps of 83.7 and 76.3 points. \textbf{The effect
reproduces in an independently sampled disjoint arm}: SK11-D9, drawn separately at 8{,}758
examples, gives 8.4\% and 28.1\%. SK11-D9 is not a seed replicate of SK12-D9b --- the two
differ in both data sample and seed --- so the agreement is across synthetic samples, not
across seeds of one sample. On the pipeline-independent metrics SK12-D9b moves \emph{upward, not downward}: comprehension
64.2\% to 67.6\%, identification 71.3\% to 73.3\%, and reference-based generation 67.9\% to
70.2\%, which are $+3.3$, $+2.0$ and $+2.3$ recovery points on the unrounded values.
\textbf{Expressed in the same unit}, the observed SK5--SK6 seed differences are $6.2$, $2.0$
and $4.9$ recovery points respectively (1.83 chrF and 1.1\,pp accuracy converted through
their own denominators; \S\ref{sec:seed}), so the three moves are 0.53, 1.00 and 0.48 times
that reference. SK11-D9's corresponding moves are $-0.4$, $+0.7$ and $+0.3$ points, all inside
the same reference. \textbf{We now also have a seed pair inside the disjoint condition.} SK7-D and SK9-D repeat
the 26{,}861-example disjoint recipe with different seeds and give held-out marker recoveries
of 14.3\%/39.0\% and 16.9\%/37.2\% --- differences of 2.6 and 1.8 points, far below the
76--84 point collapse the intervention produces. The seed reference is thus no longer borrowed
only from the coupled condition, though it is measured at 26{,}861 rather than 8{,}600 examples.

We read this as direct intervention evidence that construction access drives much of the
marker-recovery difference, rather than an evaluator swap alone --- though not a fully
isolated causal estimate, since the primary arms, matched on size, composition, register and
recipe, share only 66.2\% of their source sentences. The marker metric \emph{is} sensitive to a genuine ability
--- producing dialect lexical forms --- but its recovery estimate is strongly confounded by
whether the evaluated forms were available to the construction pipeline. \textbf{Marker-based
recovery is dominated by construction-inventory coverage and cannot by itself support claims
about broader dialect capability.}

Stated as two facts that hold simultaneously: SK12-D9b does not generalize well to dialect
lexical forms its rules never carried, recovering 8.1\% of held-out dialectness where arm R
defines 100\% (SK11-D9 gives 8.4\%); and that limitation implies no corresponding loss on
comprehension, region identification, or reference-based generation. The first is an observed
limitation of rule-based lexical generalization under this split; the second is why the marker
score cannot be interpreted as an estimate of broader dialect capability.

Finally, SK10-M holds the update budget fixed (8{,}590 steps, and therefore 8{,}590
examples at batch size one) while drawing from the 26{,}861-example pool rather than the
8{,}758-example one; it is
a \textbf{fixed-update, larger-source-pool control}, not a separation of data quantity from
optimizer steps. The larger pool raises held-out dialectness (28.1\% against 8.4\%) and
lowers identification (61.1\% against 72.0\%). We report this as a single run and draw no
directional conclusion.

\subsection{Quantifying the coupling: MCC and an overlap sweep}\label{sec:mcc}

The disjointness test is binary. To ask how coupling scales we define
\textbf{metric--construction coverage}
\[
\mathrm{MCC}=\frac{|V_{\text{eval}}\cap V_{\text{construct}}|}{|V_{\text{eval}}|},
\qquad
\mathrm{MCC}_w=\frac{\sum_{v\in V_{\text{eval}}\cap V_{\text{construct}}} f(v)}
                     {\sum_{v\in V_{\text{eval}}} f(v)},
\]
where $V_{\text{construct}}$ is the set of forms the transformation rules can emit and
$f(v)$ is the frequency of $v$ in the benchmark. Our released scoring lexicon has
$\mathrm{MCC}=1.000$ in all five regions.

\textbf{The $\mathrm{MCC}=1$ point does not reproduce the deployed evaluator's dialectness
recovery} (84.8\% against 92.3\%; the reference-chrF generation figure is 72.4\%) because the sweep holds evaluator size fixed at 1{,}168 forms per region so that
only coverage varies across levels. It is a size-matched sampled evaluator with complete
construction coverage, not the deployed evaluator; it also scores by exact match only, whereas
the deployed scorer additionally fires on suffix match, which contributes to the same gap.

To vary MCC we mine dialect word forms from the \emph{test} split, which contributed to
neither the lexicon nor the purity statistics, and which therefore contains forms outside
$V_{\text{construct}}$ (32.6--51.5\% of forms per region). Holding $|V_{\text{eval}}|$
fixed at 1{,}168 per region we vary only the fraction drawn from inside
$V_{\text{construct}}$, and rescore the stored generations. \textbf{No arm is retrained;
only the evaluator changes.}

\begin{table}[h]\centering\small
\input{tables/tab-mcc.tex}
\caption{MCC sweep on dialectness, 200 evaluator inventories per level stratified by
region only. Two uncertainties are reported
separately: ``eval.\ pct.'' is the 2.5--97.5 percentile across those evaluators, and
``item boot.'' fixes the median evaluator at that level and resamples benchmark items with
replacement (2{,}000 draws, arms paired on item index). The real-data arm stays above base
at every level under both. Endpoint recovery ranges do not overlap; adjacent levels do, so
we do not claim a difference between neighbouring rows.}\label{tab:mcc}
\end{table}

\begin{figure}[t]\centering
\includegraphics[width=3.1in]{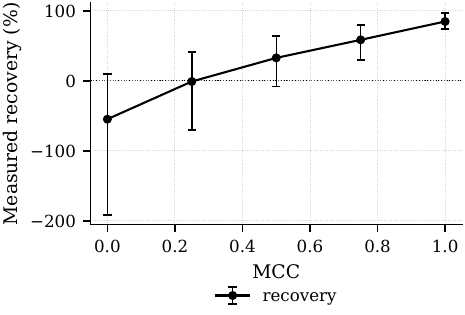}\hfill
\includegraphics[width=3.1in]{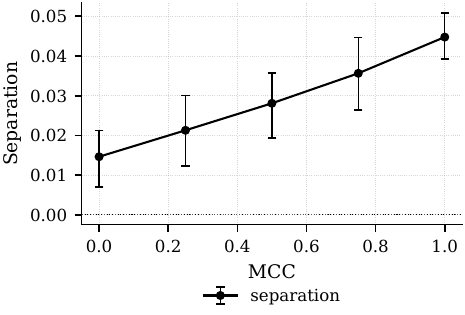}
\caption{Metric--construction coverage sweep. Models and generations are fixed; only the
evaluation inventory changes. \textbf{Left}: measured recovery of the synthetic arm (SK4)
against the real-data arm, in percent. \textbf{Right}: ``separation'' is the evaluator's own
dialectness gap between the real-data arm and base; it shrinks as coverage falls, which is
what makes the recovery ratio on the left progressively unstable --- though it never
reaches zero. Points are medians over 200 stratified evaluator samples; bars span the
2.5--97.5 percentiles.}\label{fig:mcc}
\end{figure}

To avoid confounding MCC with the identity of the sampled markers, we draw 200 evaluator
inventories at each level rather than one, stratified by region only, and report medians with the 2.5--97.5 percentile range \emph{across those
evaluator samples} (Table~\ref{tab:mcc}, Figure~\ref{fig:mcc}). We keep this interval
distinct from the paired bootstrap over benchmark items used for the arm R\,--\,base gap;
the first describes variation across evaluators, the second uncertainty in a fixed
evaluator's estimate. Measured recovery rises monotonically, from a
median of $-54.8$\% at $\mathrm{MCC}=0$ to $84.8$\% at $\mathrm{MCC}=1$, with
non-overlapping intervals at the endpoints. Nothing about the arms changed.

We keep two sources of uncertainty separate. Across the 200 evaluator inventories at a
level we report the 2.5--97.5 percentile of the statistic; for item-level uncertainty we
fix the \emph{median} evaluator at that level and resample benchmark items with
replacement (2{,}000 draws, arms paired on the item index). Under the second procedure the
arm R\,--\,base gap is above zero at every level, including $\mathrm{MCC}=0$ ($+0.0146$,
95\% CI $[+0.0084, +0.0210]$): the evaluator never goes blind. What changes is the size of the
gap, which shrinks roughly threefold from 0.0449 at $\mathrm{MCC}=1$ to 0.0146 at
$\mathrm{MCC}=0$ (item-bootstrap point estimates). Because recovery divides by that gap, its uncertainty grows sharply as
coverage falls --- the interval across evaluator samples widens from 23 points to 201. The
deployed evaluator's own gap is far from zero ($+0.1020$, CI $[+0.0949, +0.1090]$); its
problem is inflation, not insensitivity.

We also compared the two coverage statistics as predictors of the inflation, pooling all
1{,}000 sampled evaluators. Type coverage gives $r=0.841$ and frequency-weighted coverage
$r=0.854$. \textbf{We do not claim either is the better diagnostic}; a 0.013 difference in
correlation is not evidence, and we report both.

The reference metric has its own weakness: chrF against a single reference penalises
legitimate dialectal variation. We report both families rather than replacing one with
the other, and we treat any claim that rests on only the coupled family as
provisional.

\subsection{Meaning preservation is a guardrail}\label{sec:guardrail}

Meaning preservation rises as transformation decreases, so an arm that transforms less
scores higher than real data on it. It is a guardrail against degenerate copying, not
a gain axis, and it is excluded from every recovery figure we report.

\paragraph{Excluded exploratory finding.}
A marker-based over-separation pattern did not survive a marker-free probe and is excluded from our quantitative claims (Appendix~\ref{app:oversep}).

\section{Seed sensitivity}\label{sec:seed}

SK6 repeats SK5's configuration with a different seed, providing \textbf{one observed
seed-to-seed comparison under the SK5 condition}: 1.83 chrF (T1), 1.1\,pp (T2), 0.0009
dialectness, 0.0346 region match, and 4.9\,pp of reference-based generation recovery.
We use this observed difference only as an informal sensitivity reference, not as a variance estimate; taking twice the gap, SK4's advantage over SK5 is
large relative to the observed SK5--SK6 difference on identification, dialectness and region
match, and is
\emph{not} on comprehension, where the difference is a tenth of the gap; on
reference-based generation it is also within the gap. One seed pair is a single observed
comparison, not an estimate of the training procedure's variance, and the threshold has no
statistical justification --- this is a sensitivity analysis, not a significance test. The precise
statement is that increasing the synthetic training set from 8{,}600 to 26{,}600 did not produce a
comprehension improvement larger than the observed SK5--SK6 difference; we avoid the
word plateau, which would imply that other interventions would also fail, which we did not
test.

\section{Limitations}

\paragraph{A stable external evaluator is still missing.} The disjointness test in
\S\ref{sec:coupling} returns an empty held-out set for our released lexicon. Decoupled
evaluators can be built --- \S\ref{sec:mcc} constructs them, and they do separate the
real-data arm from base --- but their separation margin at $\mathrm{MCC}=0$ is roughly
three times smaller than at $\mathrm{MCC}=1$, which makes any recovery ratio computed
from them substantially less stable. We therefore still lack an externally sourced
evaluator that is both construction-independent and strongly discriminating. Building one
from a resource outside this corpus --- a published dialect dictionary, for instance --- is
the most important piece of future work this paper implies.

\paragraph{Each disjoint arm is a single training run.} The primary comparison in
\S\ref{sec:disjoint} rests on one exactly size-matched disjoint run (SK12-D9b) and one
independently sampled run at a different size (SK11-D9); neither is a seed replicate of the
other, since they differ in data sample as well as seed. The seed pair we do have inside the
disjoint condition (SK7-D/SK9-D) is at the 26{,}861-example scale, not at 8{,}600, so the
seed reference for the primary comparison is still partly borrowed. A seed replicate of
SK12-D9b would close this.

\paragraph{The arms are matched on everything we could match, but not on source sentences.}
SK12-D9b matches SK5 exactly on training-set size, task-by-region composition, source
register and optimization recipe, and shares 66.2\% of its source sentences with it. The
remaining third differs, so this is direct intervention evidence rather than a fully isolated
causal estimate. Two leaks had to be closed before the split was clean: a public dictionary
that bypassed it, and cross-region emission --- because scoring fires across regions, the ban
list has to be the union across all five rather than each region's own.

\paragraph{Over-separation is scoped to the marker axis.} The one marker-free probe we
could construct moves in the opposite direction (Appendices~\ref{app:probe} and~\ref{app:oversep}). We do not know
whether the distributional distortion is a property of the generations or of the
marker metric.

\paragraph{Two seeds, one pair.} \S\ref{sec:seed} reports a single observed seed-to-seed
comparison; it does not estimate training variance. That comparison was measured at the 8{,}600
scale under the coupled condition, and we carry it to the 26{,}600-example and disjoint arms
by assumption.

\paragraph{Data quantity and compute are confounded.} Every arm trains one epoch, so the
26{,}600-example arm also takes three times the optimizer steps; a step-matched control would separate them.

\paragraph{Both generation metrics are imperfect.} The marker family is coupled to the
pipeline; the reference family penalises legitimate variation; neither is a human judgement, and we ran none.

\paragraph{Scope.} All results are on Qwen3.8-27B; training is single-turn throughout, and
conversational ability was neither trained nor measured.

\section{Release}

We release KoDialectBench and two model checkpoints: \textbf{arm R}, the real-data reference, and \textbf{SK4}, the 26{,}600-example synthetic arm. The two arms of the primary intervention are also released as LoRA adapters with the merge script, so Figure~\ref{fig:disjoint} can be rerun against released weights (links provided upon acceptance). Because an unmerged adapter is a silent no-op on this architecture, the model cards require merging and verifier validation. Detailed checkpoint provenance and source-overlap audits are in Appendix~\ref{app:provenance}. Training sets, the marker lexicon and the glossary are withheld: the dialect-to-standard triples they encode are recoverable from the synthetic examples alone.

\bibliography{refs}

\appendix

\section{All arms, all metrics}\label{app:full}

Table~\ref{tab:full} repeats Table~\ref{tab:main} with arm C included, so the source-register
comparison in \S\ref{sec:pipeline} can be checked directly.

\begin{table}[h]\centering\scriptsize
\setlength{\tabcolsep}{3.2pt}
\input{tables/tab-full.tex}
\caption{Every arm on every benchmark metric, medians of three runs. Marker-family values
use the \emph{deployed} full lexicon; the held-out intervention-evaluator results are
reported separately in Table~\ref{tab:disjoint-result}.}\label{tab:full}
\end{table}

\section{Control preconditions and their negative tests}\label{app:controls}

The quantised-baseline confound was found by comparing recovery under a matched bf16 base against the original: the two disagreed on sign for several metrics, which is what prompted the re-measurement. The silent-adapter failure was found by comparing token log-probabilities between the base endpoint and the adapted one on identical prompts; they matched to five decimal places, which cannot happen if an adapter is active. The merge script now asserts, before an arm is served, that an adapted tensor differs from the base weights; the log-probability comparison was run by hand when the failure was found and is not an automated gate in the released code. One prompt-template discrepancy is also disclosed: the T3 training template separates instruction and input sentence by two newlines while the benchmark template uses one; the T1 and T2 templates are identical between training and evaluation.

\section{Marker-free region profile}\label{app:probe}

A character $n$-gram divergence matrix was tried first and discarded: every region pair scored 0.963--0.977, saturated, because each region's benchmark items are disjoint in content. The edit-ratio profile below is paired, so differing item sets do not matter. Human references move 0.134--0.306 depending on region.

\begin{figure}[t]\centering
\includegraphics[width=3.4in]{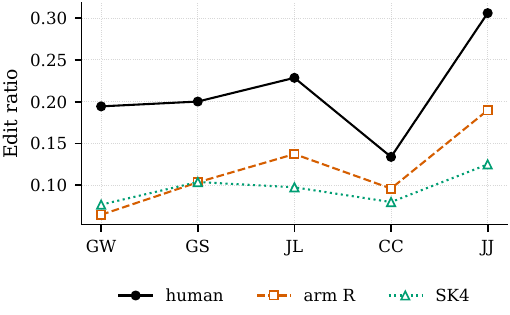}
\caption{Per-region transformation magnitude. ``Edit ratio'' is the character edit distance
between each standard input and the generated output, normalised by input length, and uses
no marker lexicon. All arms move
far less than human references, and the synthetic arm's between-region spread is narrower
rather than wider.}\label{fig:profile}
\end{figure}

\section{A small out-of-domain probe}\label{app:ood}

Benchmark items come from the same corpora arm R trained on, which favours arm R. To
probe outside that we wrote six sentences ourselves --- customer service, a public
notice, tourism, drama dialogue, game dialogue, and everyday speech --- and ran both
merged models over all five regions, using no source text ($n=30$ per model, one
greedy run).

\begin{table}[h]\centering\small
\input{tables/tab-ood.tex}
\caption{Six self-written scenes. A qualitative probe, not a benchmark.}\label{tab:ood}
\end{table}

On the formal customer-service sentence, arm R returned the identical sentence for all
five regions, ignoring the requested region. We describe this as a \emph{qualitative
reversal on region differentiation} rather than a reversal of ranking: synthetic leads
on two of three measures and trails on the third, and the third measure
(self-consistency) is the model judging its own output with its own classifier, so it
reflects internal consistency rather than correctness. The probe is too small to
support a quantitative claim, and we draw none.

\section{A second apparent finding that we exclude}\label{app:oversep}

Earlier exploratory analyses suggested that synthetic dialects separate \emph{more} sharply
between regions than real ones do on the marker-based attribution axis. \textbf{Those figures
are not ledger-backed} --- they predate our structured experiment ledger and we could not
reproduce them from the released bench records with the released scoring code --- so we
exclude them from the paper's quantitative claims rather than report them with a caveat.

A subsequent marker-free probe did not support broader over-separation. On the per-region
character edit ratio between each standard input and the generated output, every arm moves
far less than human references (arm R at 0.56 of the human magnitude, SK4 at 0.45) and the
synthetic arm's between-region spread is \emph{narrower}, not wider --- 0.31 of human against
0.76 for arm R. We report this as a failure to confirm. Whether the synthetic distribution is
distorted in some broader sense remains unresolved, and nothing in \S\S\ref{sec:axis}--\ref{sec:mcc}
depends on it. The full profile and a discarded $n$-gram probe are in
Appendix~\ref{app:probe}.

\section{Checkpoint provenance}\label{app:provenance}
On the synthetic checkpoint we state the provenance precisely. \textbf{No complete
source-corpus utterance appears verbatim in its training set}: over 11.8 million
distinct source strings, exact-match against every training sentence of six or more
whitespace-delimited eojeols (Korean spacing units) returns 0 of 26{,}600 examples, and 0 of
8{,}600 for SK12-D9b. The same script returns 8{,}600 of 8{,}600 for arm R, whose targets are
the source sentences, so the positive control is executed rather than assumed. This is not the same as containing no source-derived material. The
transformation rules are mined from source-corpus labels, and 77.2\% of transformed
words in the training set pass through those rules, with 4.6\% explained by public
resources alone. Both facts appear on the model card.

\end{document}

%% file: tables/tab-main.tex
% ⛔ 자동 생성 (make_tables.py). 손으로 고치지 마라 — 원장이 정본이다.
\begin{tabular}{lrrrrrrr}
\toprule
Metric & base & SK5 & SK6 & SK4 & SK11-D9 & SK10-M & arm R \\
\midrule
T1 chrF & 50.69 & 69.61 & 71.44 & 69.45 & 69.47 & 69.80 & \textbf{80.14} \\
T1 exact match (\%) & 2.2 & 11.7 & 11.8 & 11.6 & 10.7 & 11.0 & \textbf{24.7} \\
T2 accuracy (\%) & 24.7 & 63.7 & 64.8 & 74.6 & 64.1 & 58.1 & \textbf{79.4} \\
\midrule
T3 dialectness $^\dagger$ & 0.0337 & 0.1145 & 0.1136 & 0.1329 & 0.0822 & 0.1136 & 0.1412 \\
T3 region match $^\dagger$ & 0.1025 & 0.5889 & 0.5543 & 0.6879 & 0.5012 & 0.5847 & 0.5934 \\
T3 ADI2 $^\dagger$ & 0.0118 & 0.0852 & 0.0800 & 0.1114 & 0.0619 & 0.0840 & 0.1065 \\
T3 chrF vs.\ reference & 45.19 & 58.10 & 57.17 & 58.97 & 58.15 & 57.94 & \textbf{64.21} \\
\midrule
T3 meaning chrF $^\ddagger$ & 68.72 & 80.94 & 81.43 & 79.46 & 85.40 & 81.56 & 75.78 \\
\bottomrule
\end{tabular}

%% file: tables/tab-disjoint.tex
% ⛔ 자동 생성 (make_tables.py)
\begin{tabular}{lrrr}
\toprule
Region & scoring markers & emittable forms & held out \\
\midrule
Gangwon & 8,802 & 10,020 & 0 \\
Gyeongsang & 4,450 & 5,765 & 0 \\
Jeolla & 2,752 & 4,082 & 0 \\
Chungcheong & 1,991 & 3,263 & 0 \\
Jeju & 12,790 & 13,913 & 0 \\
\bottomrule
\end{tabular}

%% file: tables/tab-disjoint-result.tex
% ⛔ 자동 생성 (make_tables.py)
\begin{tabular}{lrrrrr}
\toprule
Arm & \multicolumn{2}{c}{marker evaluator} & \multicolumn{3}{c}{pipeline-independent} \\
\cmidrule(lr){2-3}\cmidrule(lr){4-6}
 & dialectness & region match & compr. & ident. & gen.\,ref \\
\midrule
SK5 (coupled, 8{,}600) & 91.8\% & 101.9\% & 64.2\% & 71.3\% & 67.9\% \\
\textbf{SK12-D9b} (disjoint, 8{,}600) & 8.1\% & 25.6\% & 67.6\% & 73.3\% & 70.2\% \\
SK11-D9 (disjoint, 8{,}758) & 8.4\% & 28.1\% & 63.8\% & 72.0\% & 68.1\% \\
SK10-M (disjoint pool, 8{,}590 updates) & 28.1\% & 47.5\% & 64.9\% & 61.1\% & 67.0\% \\
SK4 (coupled, 26{,}600) & 109.5\% & 122.3\% & 63.7\% & 91.2\% & 72.4\% \\
\bottomrule
\end{tabular}

%% file: tables/tab-mcc.tex
% ⛔ 자동 생성 (make_tables.py)
\begin{tabular}{rrllr@{}l}
\toprule
MCC & $\mathrm{MCC}_w$ & \multicolumn{2}{c}{arm R $-$ base} & \multicolumn{2}{c}{SK4 recovery} \\
 & & \footnotesize[eval.\ pct.] & \footnotesize[item boot.\ 95\% CI] & \multicolumn{2}{c}{\footnotesize[eval.\ 2.5--97.5 pct.]} \\
\midrule
0.00 & 0.00 & $+0.0146$ \footnotesize$[+0.0070,+0.0213]$ & $+0.0146$ \footnotesize$[+0.0084,+0.0210]$ & $-54.8$\% & $\,[-191.4,\ 9.5]$ \\
0.25 & 0.37 & $+0.0213$ \footnotesize$[+0.0123,+0.0301]$ & $+0.0213$ \footnotesize$[+0.0146,+0.0281]$ & $-0.9$\% & $\,[-70.4,\ 40.9]$ \\
0.50 & 0.64 & $+0.0281$ \footnotesize$[+0.0193,+0.0357]$ & $+0.0281$ \footnotesize$[+0.0218,+0.0343]$ & $32.8$\% & $\,[-7.9,\ 63.9]$ \\
0.75 & 0.84 & $+0.0356$ \footnotesize$[+0.0264,+0.0446]$ & $+0.0357$ \footnotesize$[+0.0301,+0.0413]$ & $58.6$\% & $\,[30.2,\ 79.6]$ \\
1.00 & 1.00 & $+0.0448$ \footnotesize$[+0.0392,+0.0508]$ & $+0.0449$ \footnotesize$[+0.0395,+0.0500]$ & $84.8$\% & $\,[73.7,\ 96.8]$ \\
\bottomrule
\end{tabular}

%% file: tables/tab-full.tex
% ⛔ 자동 생성 (make_tables.py). 손으로 고치지 마라 — 원장이 정본이다.
\begin{tabular}{lrrrrrrrrrrr}
\toprule
Metric & base & arm C & SK5 & SK6 & SK4 & SK12-D9b & SK11-D9 & SK10-M & SK7-D & SK9-D & arm R \\
\midrule
T1 chrF & 50.69 & 70.10 & 69.61 & 71.44 & 69.45 & 70.59 & 69.47 & 69.80 & 71.10 & 72.34 & \textbf{80.14} \\
T1 exact match (\%) & 2.2 & 10.6 & 11.7 & 11.8 & 11.6 & 11.7 & 10.7 & 11.0 & 13.0 & 13.6 & \textbf{24.7} \\
T2 accuracy (\%) & 24.7 & 63.1 & 63.7 & 64.8 & 74.6 & 64.8 & 64.1 & 58.1 & 71.0 & 66.8 & \textbf{79.4} \\
\midrule
T3 dialectness $^\dagger$ & 0.0337 & 0.1023 & 0.1145 & 0.1136 & 0.1329 & 0.0794 & 0.0822 & 0.1136 & 0.1030 & 0.1129 & 0.1412 \\
T3 region match $^\dagger$ & 0.1025 & 0.5094 & 0.5889 & 0.5543 & 0.6879 & 0.4785 & 0.5012 & 0.5847 & 0.6828 & 0.6862 & 0.5934 \\
T3 ADI2 $^\dagger$ & 0.0118 & 0.0653 & 0.0852 & 0.0800 & 0.1114 & 0.0584 & 0.0619 & 0.0840 & 0.0873 & 0.0972 & 0.1065 \\
T3 chrF vs.\ reference & 45.19 & 57.67 & 58.10 & 57.17 & 58.97 & 58.54 & 58.15 & 57.94 & 59.17 & 58.12 & \textbf{64.21} \\
\midrule
T3 meaning chrF $^\ddagger$ & 68.72 & 83.35 & 80.94 & 81.43 & 79.46 & 87.43 & 85.40 & 81.56 & 84.48 & 82.82 & 75.78 \\
\bottomrule
\end{tabular}

%% file: tables/tab-ood.tex
% ⛔ 자동 생성 (make_tables.py)
\begin{tabular}{lrr}
\toprule
 & arm R (real) & SK4 (synthetic) \\
\midrule
Region differentiation (5 distinct outputs) & 77\% & \textbf{97\%} \\
chrF-exact round trip & \textbf{93\%} & 83\% \\
Self-consistency of region label & 40\% & \textbf{63\%} \\
\bottomrule
\end{tabular}

%% file: main.bbl
\begin{thebibliography}{51}
\providecommand{\natexlab}[1]{#1}
\providecommand{\url}[1]{\texttt{#1}}
\expandafter\ifx\csname urlstyle\endcsname\relax
  \providecommand{\doi}[1]{doi: #1}\else
  \providecommand{\doi}{doi: \begingroup \urlstyle{rm}\Url}\fi

\bibitem[Alemohammad et~al.(2024)Alemohammad, Casco-Rodriguez, Luzi, Humayun,
  Babaei, LeJeune, Siahkoohi, and Baraniuk]{alemohammad2024mad}
Sina Alemohammad, Josue Casco-Rodriguez, Lorenzo Luzi, Ahmed~Imtiaz Humayun,
  Hossein Babaei, Daniel LeJeune, Ali Siahkoohi, and Richard~G. Baraniuk.
\newblock Self-consuming generative models go {MAD}.
\newblock In \emph{International Conference on Learning Representations
  (ICLR)}, 2024.
\newblock URL
  \url{https://proceedings.iclr.cc/paper_files/paper/2024/hash/ebc042e767de551803ccfcc45e2454f5-Abstract-Conference.html}.

\bibitem[Anikina et~al.(2025)Anikina, Cegin, Simko, and
  Ostermann]{anikina2025lowresource}
Tatiana Anikina, Jan Cegin, Jakub Simko, and Simon Ostermann.
\newblock A rigorous evaluation of {LLM} data generation strategies for
  low-resource languages.
\newblock In \emph{Proceedings of the 2025 Conference on Empirical Methods in
  Natural Language Processing}, pages 8282--8303, 2025.
\newblock URL \url{https://aclanthology.org/2025.emnlp-main.418/}.

\bibitem[Blodgett et~al.(2016)Blodgett, Green, and O'Connor]{blodgett2016aae}
Su~Lin Blodgett, Lisa Green, and Brendan O'Connor.
\newblock Demographic dialectal variation in social media: A case study of
  {A}frican-{A}merican {E}nglish.
\newblock In \emph{Proceedings of the 2016 Conference on Empirical Methods in
  Natural Language Processing}, pages 1119--1130, 2016.
\newblock URL \url{https://aclanthology.org/D16-1120/}.

\bibitem[Blodgett et~al.(2021)Blodgett, Lopez, Olteanu, Sim, and
  Wallach]{blodgett2021salmon}
Su~Lin Blodgett, Gilsinia Lopez, Alexandra Olteanu, Robert Sim, and Hanna
  Wallach.
\newblock Stereotyping {N}orwegian {S}almon: An inventory of pitfalls in
  fairness benchmark datasets.
\newblock In \emph{Proceedings of the 59th Annual Meeting of the Association
  for Computational Linguistics and the 11th International Joint Conference on
  Natural Language Processing (Volume 1: Long Papers)}, pages 1004--1015, 2021.
\newblock URL \url{https://aclanthology.org/2021.acl-long.81/}.

\bibitem[Briakou et~al.(2021)Briakou, Agrawal, Tetreault, and
  Carpuat]{briakou2021evaluating}
Eleftheria Briakou, Sweta Agrawal, Joel Tetreault, and Marine Carpuat.
\newblock Evaluating the evaluation metrics for style transfer: A case study in
  multilingual formality transfer.
\newblock In \emph{Proceedings of the 2021 Conference on Empirical Methods in
  Natural Language Processing}, pages 1321--1336, 2021.
\newblock URL \url{https://aclanthology.org/2021.emnlp-main.100/}.

\bibitem[Chim et~al.(2025)Chim, Ive, and Liakata]{chim2025synthetic}
Jenny Chim, Julia Ive, and Maria Liakata.
\newblock Evaluating synthetic data generation from user generated text.
\newblock \emph{Computational Linguistics}, 51\penalty0 (1):\penalty0 191--233,
  2025.
\newblock URL \url{https://aclanthology.org/2025.cl-1.6/}.

\bibitem[Dong et~al.(2024)Dong, Jiang, Liu, Jin, Gu, Yang, and
  Li]{dong2024generalization}
Yihong Dong, Xue Jiang, Huanyu Liu, Zhi Jin, Bin Gu, Mengfei Yang, and Ge~Li.
\newblock Generalization or memorization: Data contamination and trustworthy
  evaluation for large language models.
\newblock In \emph{Findings of the Association for Computational Linguistics:
  ACL 2024}, pages 12039--12050, 2024.
\newblock URL \url{https://aclanthology.org/2024.findings-acl.716/}.

\bibitem[Esteban et~al.(2017)Esteban, Hyland, and R{\"a}tsch]{esteban2017rcgan}
Crist{\'o}bal Esteban, Stephanie~L. Hyland, and Gunnar R{\"a}tsch.
\newblock Real-valued (medical) time series generation with recurrent
  conditional {GAN}s, 2017.
\newblock URL \url{https://arxiv.org/abs/1706.02633}.

\bibitem[Faisal et~al.(2025)Faisal, Rahman, and
  Anastasopoulos]{faisal2025toxicity}
Fahim Faisal, Md~Mushfiqur Rahman, and Antonios Anastasopoulos.
\newblock Dialectal toxicity detection: Evaluating {LLM}-as-a-judge consistency
  across language varieties.
\newblock In \emph{Findings of the Association for Computational Linguistics:
  EMNLP 2025}, pages 12429--12452, 2025.
\newblock URL \url{https://aclanthology.org/2025.findings-emnlp.664/}.

\bibitem[Fleisig et~al.(2024)Fleisig, Smith, Bossi, Rustagi, Yin, and
  Klein]{fleisig2024bias}
Eve Fleisig, Genevieve Smith, Madeline Bossi, Ishita Rustagi, Xavier Yin, and
  Dan Klein.
\newblock Linguistic bias in {C}hat{GPT}: Language models reinforce dialect
  discrimination.
\newblock In \emph{Proceedings of the 2024 Conference on Empirical Methods in
  Natural Language Processing}, pages 13541--13564, 2024.
\newblock URL \url{https://aclanthology.org/2024.emnlp-main.750/}.

\bibitem[Gill et~al.(2025)Gill, Ravichander, and Marasovic]{gill2025lost}
Alexander Gill, Abhilasha Ravichander, and Ana Marasovic.
\newblock What has been lost with synthetic evaluation?
\newblock In \emph{Findings of the Association for Computational Linguistics:
  EMNLP 2025}, pages 9902--9945, 2025.
\newblock URL \url{https://aclanthology.org/2025.findings-emnlp.526/}.

\bibitem[Groenwold et~al.(2020)Groenwold, Ou, Parekh, Honnavalli, Levy, Mirza,
  and Wang]{groenwold2020aave}
Sophie Groenwold, Lily Ou, Aesha Parekh, Samhita Honnavalli, Sharon Levy, Diba
  Mirza, and William~Yang Wang.
\newblock Investigating {A}frican-{A}merican {V}ernacular {E}nglish in
  transformer-based text generation.
\newblock In \emph{Proceedings of the 2020 Conference on Empirical Methods in
  Natural Language Processing (EMNLP)}, pages 5877--5883, 2020.
\newblock URL \url{https://aclanthology.org/2020.emnlp-main.473/}.

\bibitem[Guo et~al.(2024)Guo, Shang, Vazirgiannis, and
  Clavel]{guo2023diversity}
Yanzhu Guo, Guokan Shang, Michalis Vazirgiannis, and Chlo\'e Clavel.
\newblock The curious decline of linguistic diversity: Training language models
  on synthetic text.
\newblock In \emph{Findings of the Association for Computational Linguistics:
  NAACL 2024}, pages 3589--3604, 2024.
\newblock \doi{10.18653/v1/2024.findings-naacl.228}.
\newblock URL \url{https://aclanthology.org/2024.findings-naacl.228/}.

\bibitem[Gupta et~al.(2025)Gupta, Cheung, Meng, Sayyed, Zhu, Liao, and
  O'Brien]{gupta2025endive}
Abhay Gupta, Jacob Cheung, Philip Meng, Shayan Sayyed, Kevin Zhu, Austen Liao,
  and Sean O'Brien.
\newblock {E}n{D}ive: A cross-dialect benchmark for fairness and performance in
  large language models.
\newblock In \emph{Findings of the Association for Computational Linguistics:
  EMNLP 2025}, pages 16830--16855, 2025.
\newblock URL \url{https://aclanthology.org/2025.findings-emnlp.913/}.

\bibitem[Gururangan et~al.(2018)Gururangan, Swayamdipta, Levy, Schwartz,
  Bowman, and Smith]{gururangan2018artifacts}
Suchin Gururangan, Swabha Swayamdipta, Omer Levy, Roy Schwartz, Samuel~R.
  Bowman, and Noah~A. Smith.
\newblock Annotation artifacts in natural language inference data.
\newblock In \emph{Proceedings of the 2018 Conference of the North American
  Chapter of the Association for Computational Linguistics: Human Language
  Technologies, Volume 2 (Short Papers)}, pages 107--112, 2018.
\newblock URL \url{https://aclanthology.org/N18-2017/}.

\bibitem[Howcroft et~al.(2020)Howcroft, Belz, Clinciu, Gkatzia, Hasan,
  Mahamood, Mille, van Miltenburg, Santhanam, and
  Rieser]{howcroft2020confusion}
David~M. Howcroft, Anya Belz, Miruna-Adriana Clinciu, Dimitra Gkatzia, Sadid~A.
  Hasan, Saad Mahamood, Simon Mille, Emiel van Miltenburg, Sashank Santhanam,
  and Verena Rieser.
\newblock Twenty years of confusion in human evaluation: {NLG} needs evaluation
  sheets and standardised definitions.
\newblock In \emph{Proceedings of the 13th International Conference on Natural
  Language Generation}, pages 169--182, 2020.
\newblock URL \url{https://aclanthology.org/2020.inlg-1.23/}.

\bibitem[Hsia et~al.(2024)Hsia, Pruthi, Singh, and Lipton]{hsia2024goodhart}
Jennifer Hsia, Danish Pruthi, Aarti Singh, and Zachary~C. Lipton.
\newblock Goodhart's law applies to {NLP}'s explanation benchmarks.
\newblock In \emph{Findings of the Association for Computational Linguistics:
  EACL 2024}, pages 1322--1335, 2024.
\newblock \doi{10.18653/v1/2024.findings-eacl.88}.
\newblock URL \url{https://aclanthology.org/2024.findings-eacl.88/}.

\bibitem[Iskander et~al.(2024)Iskander, Tolmach, Shapira, Cohen, and
  Karnin]{iskander2024quality}
Shadi Iskander, Sofia Tolmach, Ori Shapira, Nachshon Cohen, and Zohar Karnin.
\newblock Quality matters: Evaluating synthetic data for tool-using {LLM}s.
\newblock In \emph{Proceedings of the 2024 Conference on Empirical Methods in
  Natural Language Processing}, pages 4958--4976, 2024.
\newblock URL \url{https://aclanthology.org/2024.emnlp-main.285/}.

\bibitem[Jacobs and Wallach(2021)]{jacobs2021measurement}
Abigail~Z. Jacobs and Hanna Wallach.
\newblock Measurement and fairness.
\newblock In \emph{Proceedings of the 2021 ACM Conference on Fairness,
  Accountability, and Transparency (FAccT '21)}, 2021.
\newblock URL \url{https://arxiv.org/abs/1912.05511}.

\bibitem[Kim et~al.(2025)Kim, Suk, Yue, Viswanathan, Lee, Wang, Gashteovski,
  Lawrence, Welleck, and Neubig]{kim2025generators}
Seungone Kim, Juyoung Suk, Xiang Yue, Vijay Viswanathan, Seongyun Lee, Yizhong
  Wang, Kiril Gashteovski, Carolin Lawrence, Sean Welleck, and Graham Neubig.
\newblock Evaluating language models as synthetic data generators.
\newblock In \emph{Proceedings of the 63rd Annual Meeting of the Association
  for Computational Linguistics (Volume 1: Long Papers)}, pages 6385--6403,
  2025.
\newblock URL \url{https://aclanthology.org/2025.acl-long.320/}.

\bibitem[Lai et~al.(2022)Lai, Mao, Toral, and Nissim]{lai2022compass}
Huiyuan Lai, Jiali Mao, Antonio Toral, and Malvina Nissim.
\newblock Human judgement as a compass to navigate automatic metrics for
  formality transfer.
\newblock In \emph{Proceedings of the 2nd Workshop on Human Evaluation of NLP
  Systems (HumEval)}, pages 102--115, 2022.
\newblock URL \url{https://aclanthology.org/2022.humeval-1.9/}.

\bibitem[Lee et~al.(2021)Lee, Kim, and Chung]{lee2021intonation}
Jooyoung Lee, Kyungwha Kim, and Minhwa Chung.
\newblock {K}orean dialect identification based on intonation modeling.
\newblock In \emph{2021 24th Conference of the Oriental COCOSDA International
  Committee for the Co-Ordination and Standardisation of Speech Databases and
  Assessment Techniques (O-COCOSDA)}, pages 168--173, 2021.
\newblock URL
  \url{https://snu.elsevierpure.com/en/publications/korean-dialect-identification-based-on-intonation-modeling}.

\bibitem[Lee et~al.(2022)Lee, Kim, and Chung]{lee2022forensic}
Jooyoung Lee, Kyung~Wha Kim, and Minhwa Chung.
\newblock {K}orean dialect identification based on an ensemble of prosodic and
  segmental feature learning for forensic speaker profiling.
\newblock In \emph{25th Conference of the Oriental COCOSDA International
  Committee for the Co-Ordination and Standardisation of Speech Databases and
  Assessment Techniques (O-COCOSDA)}, 2022.
\newblock \doi{10.1109/O-COCOSDA202257103.2022.9997971}.

\bibitem[Lee et~al.(2025)Lee, Kim, and Jeong]{lee2025wav2vec}
Ju-young Lee, Sun-hee Kim, and Min-hwa Jeong.
\newblock Automatic {K}orean dialect identification using wav2vec 2.0
  {XLS}-{R}.
\newblock \emph{Phonetics and Speech Sciences}, 17\penalty0 (3):\penalty0
  83--91, 2025.
\newblock \doi{10.13064/KSSS.2025.17.3.083}.

\bibitem[Li et~al.(2024)Li, Guo, Guerin, and Lin]{li2024contamreport}
Yucheng Li, Yunhao Guo, Frank Guerin, and Chenghua Lin.
\newblock An open-source data contamination report for large language models.
\newblock In \emph{Findings of the Association for Computational Linguistics:
  EMNLP 2024}, pages 528--541, 2024.
\newblock URL \url{https://aclanthology.org/2024.findings-emnlp.30/}.

\bibitem[Lin et~al.(2025)Lin, Mao, La~Malfa, Hofmann, de~Wynter, Wang, Chen,
  Wooldridge, Pierrehumbert, and Wei]{lin2025redial}
Fangru Lin, Shaoguang Mao, Emanuele La~Malfa, Valentin Hofmann, Adrian
  de~Wynter, Xun Wang, Si-Qing Chen, Michael~J. Wooldridge, Janet~B.
  Pierrehumbert, and Furu Wei.
\newblock Assessing dialect fairness and robustness of large language models in
  reasoning tasks.
\newblock In \emph{Proceedings of the 63rd Annual Meeting of the Association
  for Computational Linguistics (Volume 1: Long Papers)}, pages 6317--6342,
  2025.
\newblock URL \url{https://aclanthology.org/2025.acl-long.317/}.

\bibitem[Liu et~al.(2023)Liu, Held, and Yang]{liu2023dada}
Yanchen Liu, William Held, and Diyi Yang.
\newblock {DADA}: Dialect adaptation via dynamic aggregation of linguistic
  rules.
\newblock In \emph{Proceedings of the 2023 Conference on Empirical Methods in
  Natural Language Processing}, pages 13776--13793, 2023.
\newblock URL \url{https://aclanthology.org/2023.emnlp-main.850/}.

\bibitem[Long et~al.(2024)Long, Wang, Xiao, Zhao, Ding, Chen, and
  Wang]{long2024survey}
Lin Long, Rui Wang, Ruixuan Xiao, Junbo Zhao, Xiao Ding, Gang Chen, and Haobo
  Wang.
\newblock On {LLM}s-driven synthetic data generation, curation, and evaluation:
  A survey.
\newblock In \emph{Findings of the Association for Computational Linguistics:
  ACL 2024}, pages 11065--11082, 2024.
\newblock URL \url{https://aclanthology.org/2024.findings-acl.658/}.

\bibitem[Maheshwari et~al.(2026)Maheshwari, Rajpoot, Cocarascu, and
  {Mamta}]{maheshwari2026lora}
Sanjh Maheshwari, Aniket~Singh Rajpoot, Oana Cocarascu, and {Mamta}.
\newblock Improving dialect robustness in large language models via {L}o{RA}
  and mixture-of-experts.
\newblock In \emph{Proceedings of the 13th Workshop on NLP for Similar
  Languages, Varieties and Dialects}, pages 293--303, 2026.
\newblock URL \url{https://aclanthology.org/2026.vardial-1.24/}.

\bibitem[Manheim and Garrabrant(2018)]{manheim2018goodhart}
David Manheim and Scott Garrabrant.
\newblock Categorizing variants of goodhart's law, 2018.
\newblock URL \url{https://arxiv.org/abs/1803.04585}.
\newblock Preprint; no archival venue.

\bibitem[Mir et~al.(2019)Mir, Felbo, Obradovich, and Rahwan]{mir2019evaluating}
Remi Mir, Bjarke Felbo, Nick Obradovich, and Iyad Rahwan.
\newblock Evaluating style transfer for text.
\newblock In \emph{Proceedings of the 2019 Conference of the North American
  Chapter of the Association for Computational Linguistics: Human Language
  Technologies, Volume 1 (Long and Short Papers)}, pages 495--504, 2019.
\newblock URL \url{https://aclanthology.org/N19-1049/}.

\bibitem[Mukherjee et~al.(2025)Mukherjee, Ojha, McCrae, and
  Du{\v{s}}ek]{mukherjee2025reliable}
Sourabrata Mukherjee, Atul~Kr. Ojha, John~P. McCrae, and Ond{\v{r}}ej
  Du{\v{s}}ek.
\newblock Evaluating text style transfer evaluation: Are there any reliable
  metrics?
\newblock In \emph{Proceedings of the 2025 Conference of the Nations of the
  Americas Chapter of the Association for Computational Linguistics: Human
  Language Technologies (Volume 4: Student Research Workshop)}, pages 418--434,
  2025.
\newblock URL \url{https://aclanthology.org/2025.naacl-srw.41/}.

\bibitem[{National Information Society Agency (NIA)}(2020)]{aihub_dialect}
{National Information Society Agency (NIA)}.
\newblock {H}an'gugeo {B}angeon {B}alhwa [{K}orean {D}ialectal {S}peech],
  regional corpora for {G}angwon, {G}yeongsang, {J}eolla, {J}eju and
  {C}hungcheong; and {J}ung-{N}onyeoncheung {H}an'gugeo {B}angeon {D}eiteo
  [{K}orean {D}ialect {D}ata for {M}iddle-aged and {E}lderly {S}peakers].
\newblock AI Hub,
  \url{https://aihub.or.kr/aihubdata/data/view.do?dataSetSn=118}, 2020.
\newblock Datasets 118--122 (constructed 2020) and 71517, 71558. AI Hub
  publishes no official English dataset title; the bracketed glosses are ours.
  Redistribution and export restricted, which is why this paper releases
  identifier hashes rather than items.

\bibitem[Ostheimer et~al.(2023)Ostheimer, Nagda, Kloft, and
  Fellenz]{ostheimer2023call}
Phil Ostheimer, Mayank Nagda, Marius Kloft, and Sophie Fellenz.
\newblock A call for standardization and validation of text style transfer
  evaluation.
\newblock In \emph{Findings of the Association for Computational Linguistics:
  ACL 2023}, pages 10791--10815, 2023.
\newblock URL \url{https://aclanthology.org/2023.findings-acl.687/}.

\bibitem[Ostheimer et~al.(2024)Ostheimer, Nagda, Kloft, and
  Fellenz]{ostheimer2024llm}
Phil Ostheimer, Mayank Nagda, Marius Kloft, and Sophie Fellenz.
\newblock Text style transfer evaluation using large language models.
\newblock In \emph{Proceedings of the 2024 Joint International Conference on
  Computational Linguistics, Language Resources and Evaluation (LREC-COLING
  2024)}, pages 15802--15822, 2024.
\newblock URL \url{https://aclanthology.org/2024.lrec-main.1373/}.

\bibitem[Park et~al.(2026)Park, Yu, and Kim]{park2026korean}
Keunhyeung Park, Seunguk Yu, and Youngbin Kim.
\newblock Steering {LLM}s toward {K}orean {L}ocal {S}peech: Iterative
  refinement framework for {F}aithful {D}ialect {T}ranslation.
\newblock In \emph{Proceedings of the Fifteenth Language Resources and
  Evaluation Conference}, pages 3257--3269, May 2026.
\newblock \doi{10.63317/2g2idiqbj4xj}.
\newblock URL \url{https://aclanthology.org/2026.lrec-1.256/}.

\bibitem[Pauli et~al.(2025)Pauli, Augenstein, and Assent]{pauli2025stylegap}
Amalie~Brogaard Pauli, Isabelle Augenstein, and Ira Assent.
\newblock Mind the style gap: Meta-evaluation of style and attribute transfer
  metrics.
\newblock In \emph{Findings of the Association for Computational Linguistics:
  EMNLP 2025}, pages 21550--21564, 2025.
\newblock URL \url{https://aclanthology.org/2025.findings-emnlp.1175/}.

\bibitem[Poliak et~al.(2018)Poliak, Naradowsky, Haldar, Rudinger, and
  Van~Durme]{poliak2018hypothesis}
Adam Poliak, Jason Naradowsky, Aparajita Haldar, Rachel Rudinger, and Benjamin
  Van~Durme.
\newblock Hypothesis only baselines in natural language inference.
\newblock In \emph{Proceedings of the Seventh Joint Conference on Lexical and
  Computational Semantics}, pages 180--191, 2018.
\newblock URL \url{https://aclanthology.org/S18-2023/}.

\bibitem[Popovi{\'c}(2015)]{popovic2015chrf}
Maja Popovi{\'c}.
\newblock chr{F}: character n-gram {F}-score for automatic {MT} evaluation.
\newblock In \emph{Proceedings of the Tenth Workshop on Statistical Machine
  Translation}, pages 392--395, 2015.
\newblock \doi{10.18653/v1/W15-3049}.
\newblock URL \url{https://aclanthology.org/W15-3049/}.

\bibitem[Ribeiro et~al.(2020)Ribeiro, Wu, Guestrin, and
  Singh]{ribeiro2020checklist}
Marco~Tulio Ribeiro, Tongshuang Wu, Carlos Guestrin, and Sameer Singh.
\newblock Beyond accuracy: Behavioral testing of {NLP} models with
  {C}heck{L}ist.
\newblock In \emph{Proceedings of the 58th Annual Meeting of the Association
  for Computational Linguistics}, pages 4902--4912, 2020.
\newblock URL \url{https://aclanthology.org/2020.acl-main.442/}.

\bibitem[Sainz et~al.(2023)Sainz, Campos, Garc{\'i}a-Ferrero, Etxaniz,
  de~Lacalle, and Agirre]{sainz2023trouble}
Oscar Sainz, Jon Campos, Iker Garc{\'i}a-Ferrero, Julen Etxaniz, Oier~Lopez
  de~Lacalle, and Eneko Agirre.
\newblock {NLP} evaluation in trouble: On the need to measure {LLM} data
  contamination for each benchmark.
\newblock In \emph{Findings of the Association for Computational Linguistics:
  EMNLP 2023}, pages 10776--10787, 2023.
\newblock URL \url{https://aclanthology.org/2023.findings-emnlp.722/}.

\bibitem[Sainz et~al.(2024)Sainz, Garc{\'i}a-Ferrero, Jacovi, Campos, Elazar,
  Agirre, Goldberg, Chen, Chim, Choshen, D'Amico-Wong, Dell, Fan, Golchin, Li,
  Liu, Pahwa, Prabhu, Sharma, Silcock, Solonko, Stap, Surdeanu, Tseng,
  Udandarao, Wang, Xu, and Yang]{sainz2024conda}
Oscar Sainz, Iker Garc{\'i}a-Ferrero, Alon Jacovi, Jon~Ander Campos, Yanai
  Elazar, Eneko Agirre, Yoav Goldberg, Wei-Lin Chen, Jenny Chim, Leshem
  Choshen, Luca D'Amico-Wong, Melissa Dell, Run-Ze Fan, Shahriar Golchin,
  Yucheng Li, Pengfei Liu, Bhavish Pahwa, Ameya Prabhu, Suryansh Sharma, Emily
  Silcock, Kateryna Solonko, David Stap, Mihai Surdeanu, Yu-Min Tseng, Vishaal
  Udandarao, Zengzhi Wang, Ruijie Xu, and Jinglin Yang.
\newblock Data contamination report from the 2024 {CONDA} shared task.
\newblock In \emph{Proceedings of the 1st Workshop on Data Contamination
  (CONDA)}, pages 41--56, 2024.
\newblock \doi{10.18653/v1/2024.conda-1.4}.
\newblock URL \url{https://aclanthology.org/2024.conda-1.4/}.

\bibitem[Santill{\'a}n~Cooper et~al.(2025)Santill{\'a}n~Cooper, Ashktorab, Do,
  Miehling, Geyer, Gajcin, Daly, Pan, and Desmond]{cooper2025evalassist}
Mart{\'i}n Santill{\'a}n~Cooper, Zahra Ashktorab, Hyo~Jin Do, Erik Miehling,
  Werner Geyer, Jasmina Gajcin, Elizabeth~M. Daly, Qian Pan, and Michael
  Desmond.
\newblock Synthetic data for evaluation: Supporting {LLM}-as-a-judge workflows
  with {E}val{A}ssist.
\newblock In \emph{Proceedings of the 2025 Conference on Empirical Methods in
  Natural Language Processing: System Demonstrations}, pages 1--11, 2025.
\newblock URL \url{https://aclanthology.org/2025.emnlp-demos.1/}.

\bibitem[Schmidtova et~al.(2024)Schmidtova, Mahamood, Balloccu, Dusek, Gatt,
  Gkatzia, Howcroft, Platek, and Sivaprasad]{schmidtova2024metrics}
Patricia Schmidtova, Saad Mahamood, Simone Balloccu, Ondrej Dusek, Albert Gatt,
  Dimitra Gkatzia, David~M. Howcroft, Ondrej Platek, and Adarsa Sivaprasad.
\newblock Automatic metrics in natural language generation: A survey of current
  evaluation practices.
\newblock In \emph{Proceedings of the 17th International Natural Language
  Generation Conference}, pages 557--583, 2024.
\newblock URL \url{https://aclanthology.org/2024.inlg-main.44/}.

\bibitem[Shumailov et~al.(2024)Shumailov, Shumaylov, Zhao, Papernot, Anderson,
  and Gal]{shumailov2024collapse}
Ilia Shumailov, Zakhar Shumaylov, Yiren Zhao, Nicolas Papernot, Ross Anderson,
  and Yarin Gal.
\newblock {AI} models collapse when trained on recursively generated data.
\newblock \emph{Nature}, 631:\penalty0 755--759, 2024.
\newblock \doi{10.1038/s41586-024-07566-y}.
\newblock URL \url{https://doi.org/10.1038/s41586-024-07566-y}.

\bibitem[Tikhonov et~al.(2019)Tikhonov, Shibaev, Nagaev, Nugmanova, and
  Yamshchikov]{tikhonov2019retrain}
Alexey Tikhonov, Viacheslav Shibaev, Aleksander Nagaev, Aigul Nugmanova, and
  Ivan~P. Yamshchikov.
\newblock Style transfer for texts: Retrain, report errors, compare with
  rewrites.
\newblock In \emph{Proceedings of the 2019 Conference on Empirical Methods in
  Natural Language Processing and the 9th International Joint Conference on
  Natural Language Processing (EMNLP-IJCNLP)}, pages 3936--3945, 2019.
\newblock URL \url{https://aclanthology.org/D19-1406/}.

\bibitem[van~der Lee et~al.(2019)van~der Lee, Gatt, van Miltenburg, Wubben, and
  Krahmer]{vanderlee2019best}
Chris van~der Lee, Albert Gatt, Emiel van Miltenburg, Sander Wubben, and Emiel
  Krahmer.
\newblock Best practices for the human evaluation of automatically generated
  text.
\newblock In \emph{Proceedings of the 12th International Conference on Natural
  Language Generation}, pages 355--368, 2019.
\newblock URL \url{https://aclanthology.org/W19-8643/}.

\bibitem[Yakhni and Chehab(2025)]{yakhni2025lebanese}
Silvana Yakhni and Ali Chehab.
\newblock Fine-tuning {LLM}s for low-resource dialect translation: The case of
  {L}ebanese, 2025.

\bibitem[Zhu et~al.(2024)Zhu, Cheng, Peng, Li, Peng, Liu, Qiu, and
  Huang]{zhu2024decontamination}
Qin Zhu, Qinyuan Cheng, Runyu Peng, Xiaonan Li, Ru~Peng, Tengxiao Liu, Xipeng
  Qiu, and Xuanjing Huang.
\newblock Inference-time decontamination: Reusing leaked benchmarks for large
  language model evaluation.
\newblock In \emph{Findings of the Association for Computational Linguistics:
  EMNLP 2024}, pages 9113--9129, 2024.
\newblock URL \url{https://aclanthology.org/2024.findings-emnlp.532/}.

\bibitem[Ziems et~al.(2022)Ziems, Chen, Harris, Anderson, and
  Yang]{ziems2022value}
Caleb Ziems, Jiaao Chen, Camille Harris, Jessica Anderson, and Diyi Yang.
\newblock {VALUE}: Understanding dialect disparity in {NLU}.
\newblock In \emph{Proceedings of the 60th Annual Meeting of the Association
  for Computational Linguistics (Volume 1: Long Papers)}, pages 3701--3720,
  2022.
\newblock URL \url{https://aclanthology.org/2022.acl-long.258/}.

\bibitem[Ziems et~al.(2023)Ziems, Held, Yang, Dhamala, Gupta, and
  Yang]{ziems2023multivalue}
Caleb Ziems, William Held, Jingfeng Yang, Jwala Dhamala, Rahul Gupta, and Diyi
  Yang.
\newblock Multi-{VALUE}: A framework for cross-dialectal {E}nglish {NLP}.
\newblock In \emph{Proceedings of the 61st Annual Meeting of the Association
  for Computational Linguistics (Volume 1: Long Papers)}, pages 744--768, 2023.
\newblock URL \url{https://aclanthology.org/2023.acl-long.44/}.

\end{thebibliography}
